\documentclass[aps,pre,twocolumn,showpacs,superscriptaddress]{revtex4-2}

\usepackage{amsmath,amssymb,amsfonts}
\usepackage{bm}
\usepackage{hyperref}
\usepackage{xcolor}
\usepackage{subcaption}
\usepackage{graphicx}
\usepackage{placeins}

\graphicspath{{figures/}{../lnn-advanced/src/}{../final_N20_trajectories/}{../final_4_special_trajectories/}{../}}

\newcommand{\AT}{V_{\mathrm{AT}}}
\newcommand{\LJ}{V_{\mathrm{LJ}}}
\newcommand{\angfac}{(1 + 3\cos\theta_1\cos\theta_2\cos\theta_3)}

\begin{document}

\title{Recovering Irreducible Three-Body Interactions with Lagrangian Neural Networks}

\author{Sena Kalabal\i k}
\email{sena.kalabalik@std.bogazici.edu.tr}
\affiliation{Department of Physics, Bo\u{g}azi\c{c}i University, Istanbul, T\"urkiye}

\author{\c{S}\"ukr\"u \c{C}a\u{g}lar}
\affiliation{Department of Physics, Bo\u{g}azi\c{c}i University, Istanbul, T\"urkiye}

\author{Abdullah Umut Hamzao\u{g}ullar\i}
\affiliation{Department of Physics, Bo\u{g}azi\c{c}i University, Istanbul, T\"urkiye}
\affiliation{Department of Computer Engineering, Bo\u{g}azi\c{c}i University, Istanbul, T\"urkiye}

\author{Arkada\c{s} \"Ozak\i n}
\affiliation{Department of Physics, Bo\u{g}azi\c{c}i University, Istanbul, T\"urkiye}

\date{\today}

\begin{abstract}
Lagrangian Neural Networks (LNNs) learn mechanical systems from trajectory data by parameterizing a scalar Lagrangian and deriving the dynamics through the Euler-Lagrange equations, biasing the model toward motion that respects the variational structure of mechanics. The interactions studied so far have been pairwise-additive, leaving open whether LNNs can recover an irreducible many-body interaction from motion alone. We address this using the Axilrod-Teller triple-dipole interaction, the leading three-body dispersion term, whose dependence on the collective geometry of three particles cannot be represented by any sum of pairwise terms. We prescribe the kinetic energy exactly and learn only the potential, as a function of interparticle distances and angles. Combined with a sampling strategy that concentrates training data where the three-body signal is strongest while preserving broad coverage of configuration space, this recovers the interaction from accelerations alone. The learned potential reproduces the radial scaling, angular dependence, and characteristic sign reversal of the Axilrod-Teller interaction, and generates stable trajectories. Embedding it within a pairwise Lennard-Jones potential lets us examine its observability: the three-body term is recovered once its dynamical signature exceeds the error of the pairwise fit, which it does cleanly at amplified coupling but not at the coupling of a real noble gas, bracketing the transition between the two. Supplying the pairwise term exactly makes it recoverable at the physical coupling too, showing that the obstruction is not the three-body term itself but its small size relative to the pairwise interaction. Finally, symbolic regression on the trained potential, with no assumed functional form, returns the Axilrod-Teller expression with angular numerator, radial law, and coupling constant intact. An LNN with domain-specific structure can thus recover an irreducible three-body interaction from trajectories as a closed-form law, pointing toward learning interaction laws whose structure is not known a priori.
\end{abstract}

\maketitle

\section{Introduction}
\label{sec:intro}

Lagrangian Neural Networks (LNNs) are physics-informed architectures that incorporate the principle of stationary action into the model structure. In a standard LNN \cite{ Cranmer2020}, a single neural network parameterizes the scalar Lagrangian \(L(q,\dot{q})\), while the system dynamics are obtained by differentiating this function through the Euler--Lagrange equations. This construction provides an inductive bias toward dynamics that are consistent with a variational formulation of mechanics, instead of enforcing such structure through external loss terms or constraints.

The systems on which these architectures have been developed and tested do include cases that are chaotic, highly nonlinear, and well described by Lagrangian mechanics, such as pendula, oscillators, gravitational pairs \cite{Cranmer2020,Lutter2019,Finzi2020, hamza2026}. However, to our knowledge, non-additive many-body interactions, potentials that cannot be written as a sum over pairs at all, have not been the object of such a study. A non-additive three-body potential changes the learning problem qualitatively, because the network must represent irreducible correlations among all interparticle coordinates simultaneously, and no composition of pairwise fits can reproduce it.

We therefore focus on an irreducible three-body interaction: the Axilrod--Teller (AT) triple-dipole term, derived independently by Axilrod and Teller \cite{Axilrod1943} and by Muto \cite{Muto1943} as the direct quantum-mechanical result, at third order in perturbation theory, for three neutral atoms. In practical atomistic models, it is commonly incorporated as the leading three-body correction to pairwise interaction potentials such as the Lennard--Jones (LJ) potential \cite{Sadus1998, Polymeropoulos1984}. The interaction has certain analytic virtues: its radial and angular structure are known in closed form, and its geometric content, which is an angular factor that changes sign with the shape of the particle triangle, is a good study case for the kind of many-body structure that a pairwise model is provably unable to represent. We analyze the nature of the interaction in detail and use its functional form as a solid target for training and testing LNNs in the context of these three-body interactions. 

Our strategy is to incorporate into the LNN framework the physics that is known a priori, leaving the network to learn only what cannot be
specified in advance. As in the general LNN settings, learner sees only trajectories, and the potential is not supervised directly, it is inferred implicitly through acceleration matching. Our experiments evaluate this program in the context of learnability at three levels: field-level accuracy, functional geometric structure, and trajectory rollouts  for the isolated AT system, and for additional systems including a purely pairwise Lennard-Jones control, and AT embedded in a Lennard-Jones background, where we investigate the coupling strength at which the three-body correction becomes identifiable at all. 

Finally, we ask an additional question. If the network succeeds in learning the interaction from trajectories alone, can the learned function be turned back into an explicit analytic expression? We therefore also subject the trained model to symbolic regression using the learned potential, turning the neural representation into a candidate functional form. This is relevant for the broader goal of using learned dynamical models as tools for scientific discovery: numerical agreement with trajectories establishes that a model works, and recovering a compact analytic expression can reveal what interaction the model has actually learned. In our study, the symbolic model recovers the characteristic structure of the interaction, including its geometric dependence, providing a clear example of how physics-informed neural learning can be paired with symbolic methods to move from dynamical data toward interpretable analytic forms.

The remainder of this paper is organized as follows. Section~\ref{sec:background} introduces the Lagrangian neural network framework and the physical systems considered, and reviews Lagrangian--Hamiltonian learning architectures, machine-learned interatomic potentials, and the Axilrod--Teller interaction. Section~\ref{sec:system-def} specifies the physical system and its interaction model. Section~\ref{sec:methods} presents our fixed-kinetic architecture, augmented input representation, and training methodology. Sections~\ref{sec:exp_at}--\ref{sec:exp_ljat} evaluate the learned models across the isolated AT, pairwise LJ, and combined LJ--AT systems, and Sec.~\ref{sec:exp_at_symbolic} presents the symbolic-regression analysis of the learned interaction. We conclude in Sec.~\ref{sec:discussion}.

\section{Background and Related Work}
\label{sec:background}

\subsection{Lagrangian neural networks}
\label{sec:lnn-background}

Consider a mechanical system with generalized coordinates $\bm q$, velocities $\dot{\bm q}$, and Lagrangian $\mathcal{L}(\bm q,\dot{\bm q})$. The Euler--Lagrange equations yield the accelerations
\begin{equation}
  \ddot{\bm q}
  = M^{-1}
    \left[\nabla_{\bm q}\mathcal{L}
    - \left(\nabla_{\bm q}\nabla_{\dot{\bm q}}^{\top}\mathcal{L}\right)\dot{\bm q}\right],
  \qquad
  M \equiv \nabla_{\dot{\bm q}}\nabla_{\dot{\bm q}}^{\top}\mathcal{L},
  \label{eq:EL}
\end{equation}
where $M$ is the generalized mass matrix. A Lagrangian neural network (LNN) represents $\mathcal{L}$ by a neural network and evaluates Eq.~\eqref{eq:EL} by automatic differentiation \cite{Cranmer2020}. Training uses trajectory or state data and differentiates the learned Lagrangian to obtain accelerations through the Euler--Lagrange equations, and compares them against reference accelerations.

\subsection{Learning Lagrangian and Hamiltonian dynamics}
\label{sec:related-lnn}

Hamiltonian neural networks \cite{Greydanus2019} and Lagrangian neural networks \cite{Cranmer2020} parameterize a scalar generating function and obtain dynamics by differentiation, thereby biasing the learned dynamics toward conservation laws through their structure, instead of supervised learning of the dynamics. The standard LNN leaves the entire $\mathcal{L}_\theta(\bm q,\dot{\bm q})$ free, maximally flexible, with the cost of the learned kinetic structure being a fitting burden of its own. Deep Lagrangian Networks (DeLaN) \cite{Lutter2019} impose a different structure than LNN, parameterizing a positive-definite mass matrix $M_\theta(\bm q)$ and a potential $V_\theta(\bm q)$ separately. Finzi \emph{et al.} \cite{Finzi2020} showed that embedding classical mechanical systems in Cartesian coordinates with explicit constraints reduces the mass matrix to a constant, few-parameter form, specified directly, hence often simplifying the learned Lagrangian and reducing the learning task to the potential energy. More recently, Hamzao\u{g}ullar\i{} \& \"Ozak\i n \cite{hamza2026} retained a fully learned Lagrangian while introducing Hessian regularization to improve optimization stability. Finally, graph-network formulations extend Hamiltonian and Lagrangian learning to many-body and articulated systems, generalizing across system size by construction \cite{Sanchez2019, Bhattoo2021, Bhattoo2022a, Bhattoo2022b}, and more recent work couples such architectures with symbolic regression to recover pairwise interaction laws directly from trajectories, including for complex potentials such as binary Lennard-Jones \cite{Bishnoi2023}. However, these formulations parameterize interactions through edge-based or pairwise decomposition rather than explicit higher-order terms, leaving irreducible three-body contributions such as those arising from the Axilrod-Teller potential unaddressed.

\subsection{The Axilrod--Teller interaction}
\label{sec:theory}

At third order in perturbation theory, the dipole--dipole interaction among three polarizable atoms produces an energy contribution that involves all three simultaneously. Applying the London approximation to the sums over intermediate states and retaining only the dipole--dipole contribution reduces the third-order perturbation expression to the Axilrod--Teller--Muto triple-dipole potential:
\cite{Axilrod1943,Stone2013}
\begin{equation}
  \AT = \nu\,\frac{\angfac}{(r_{12}\,r_{23}\,r_{31})^3},
  \label{eq:AT}
\end{equation}
where $r_{ij}$ are the side lengths of the triangle formed by the atoms, $\theta_i$ its interior angles, and $\nu>0$ a coupling constant built from atomic polarizabilities and excitation energies.

Two structural features of Eq.~\eqref{eq:AT} organize everything that follows. First, under a uniform rescaling of the triangle, $r_{ij}\to\lambda r_{ij}$, the angles are unchanged and $\AT\to\lambda^{-9}\AT$: every fixed shape obeys the same $r^{-9}$ power law. Second, the angular factor $\Phi \equiv \angfac$ changes sign at $\cos\theta_1\cos\theta_2\cos\theta_3=-1/3$ and is bounded, $\Phi\in[-2,\,11/8]$, with the maximum at the equilateral configuration and the minimum in the collinear limit. For $\nu>0$, the interaction is therefore repulsive for compact, near-equilateral geometries and attractive for elongated, near-collinear ones. This sign change is a distinctive three-body feature and provides a qualitative signature that a learned potential should reproduce to capture physics beyond a pairwise fit.

\begin{figure}[t]
    \centering
    \includegraphics[width=0.95\linewidth]{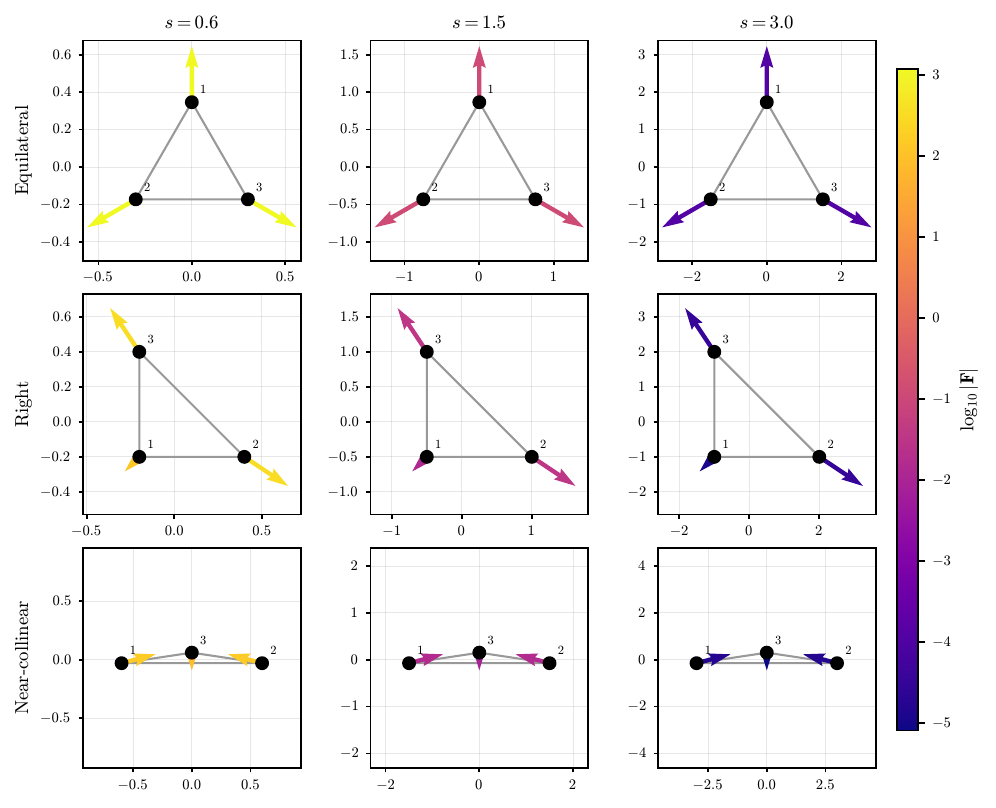}
    \caption{The AT interaction at a glance: analytic forces on all three particles for representative triangle shapes (rows) at increasing size (columns), with arrow color encoding force magnitude. Compact, near-equilateral shapes ($\Phi>0$) are pushed apart; elongated, near-collinear shapes ($\Phi<0$) are pulled together; the overall strength falls dramatically as $r^{-10}$ in force.}
    \label{fig:at_schematic}
\end{figure}

Fig.~\ref{fig:at_schematic} shows these two features directly, as forces on concrete configurations, before any learning enters the discussion. We note also that the natural variables of Eq.~\eqref{eq:AT} are exactly the six scalars $\{r_{12},r_{23},r_{31},\cos\theta_1,\cos\theta_2,\cos\theta_3\}$: this observation is relevant to our network's input representation in Sec.~\ref{sec:input-rep}.

\subsection{Machine-learned interatomic potentials}
\label{sec:related-mlp}

A substantial literature learns interatomic potentials from quantum-mechanical reference data, from the symmetry-function networks of Behler and Parrinello \cite{Behler2007,Behler2011}, which represent local atomic environments through functions of interatomic distances and angles, to neural network potentials trained at scale \cite{Smith2017} and equivariant message-passing architectures \cite{Batzner2022}; see Ref.~\cite{Kocer2022} for a concise overview, and Ref.~\cite{Kang2023} for examples incorporating explicit many-body energy terms. These models are typically trained against energies, often together with forces, computed from electronic-structure calculations. As such, the supervision reflects the combined effect of all interactions present in the underlying Hamiltonian simultaneously, not intending to identify or recover any single physical contribution in isolation. Kernel-based approaches have targeted non-additive three-body structure, from many-body distribution functionals -including AT- \cite{Khan2023} to three-body force fields for molecular liquids \cite{Scherer2020}, but again through regression on the quantum-mechanical energetics, not on the dynamical trajectories.

Our setting differs in two respects. First, we consider a single isolated interaction, allowing recovery of a specific physical contribution. Second, the potential is never supervised directly, it is recovered implicitly from classical trajectories through acceleration matching, without energy or potential labels. Our input representation (pairwise distances augmented with angle cosines, Sec.~\ref{sec:input-rep}) follows a similar philosophy to Behler symmetry functions, enriching pairwise distance information with angular information while avoiding engineered symmetry-function expansions.

\subsection{The Axilrod--Teller interaction in simulation}
\label{sec:related-at}

The AT term has a long history in rare-gas simulation. Barker, Fisher, and Watts incorporated it into quantitative models of liquid argon \cite{Barker1971}, while Attard's simulation and integral-equation studies of the LJ$+$AT fluid \cite{Attard1992a,Attard1992b} established the argon-like parameter conventions ($\tilde\nu\approx0.1$) that we adopt for physical orientation, under which the three-body interaction increases liquid-state pressure and internal energy by roughly ten percent under representative liquid-state conditions. Its physical consequences extend from crystal-structure selection in noble-gas solids \cite{Barker1971} to cohesive-energy corrections in dense helium \cite{Loubeyre1987}, making it the canonical three-body contribution to the many-body dispersion physics incorporated in modern electronic-structure methods \cite{Tkatchenko2012}. Equally relevant to the present work is the literature on the AT form's limitations: it is an asymptotic long-range interaction, and at sufficiently short separations three-body exchange interactions become comparable to and eventually exceed the triple-dipole contribution \cite{Polymeropoulos1984}. Consequently, even simulations that treat the bare AT term as the sole three-body interaction, including phase-coexistence studies such as Ref.~\cite{Sadus1998}, inherit this short-range limitation. These limits are reflected in the explicit statement of scope given in Sec.~\ref{sec:system-def}.

\section{Physical System Definition}
\label{sec:system-def}

We consider three identical particles of mass $m$ in two spatial dimensions with total potential
\begin{equation}
  V = \sum_{i<j} \LJ(r_{ij}) + \AT,
  \qquad
  \LJ(r) = 4\epsilon\!\left[\Big(\frac{\sigma}{r}\Big)^{12}-\Big(\frac{\sigma}{r}\Big)^{6}\right].
  \label{eq:total-V}
\end{equation}
We express all dynamical quantities in reduced LJ units, $r^*=r/\sigma$, $V^*=V/\epsilon$, $t^*=t\sqrt{\epsilon/m\sigma^2}$, in which the only free parameter of the dynamics is the dimensionless three-body coupling $\tilde\nu=\nu/\epsilon\sigma^9$. Hence in these units, 

\begin{equation}
V^* =
\sum_{i<j}4\left[(r_{ij}^*)^{-12}-(r_{ij}^*)^{-6}\right]
+\tilde{\nu}\,
\frac{1+3\cos\theta_1\cos\theta_2\cos\theta_3}
{(r_{12}^*r_{23}^*r_{31}^*)^3},
\end{equation}

For notational convenience, we suppress the stars on all quantities below. Unless otherwise stated, all lengths, energies, times, and coordinates are therefore understood to be expressed in these reduced LJ units.

Argon-like values ($\sigma=3.405$~\AA, $\epsilon/k_B=120$~K, $\tilde\nu\approx0.1$ \cite{Attard1992a,Attard1992b}) serve only to interpret parameter regimes physically, the dynamics are solved in reduced units throughout. The isolated three-body system of Sec.~\ref{sec:exp_at} corresponds to $\epsilon_{\mathrm{LJ}}=0$ with $\nu$ setting the (then arbitrary) energy scale. 

The interaction model is directly inspired by classical many-body fluid Hamiltonians incorporating both pairwise Lennard-Jones and three-body Axilrod--Teller contributions: the functional form of the triple-dipole term is identical to that introduced in early Monte Carlo studies of rare-gas systems, where the Hamiltonian is expressed as a sum over two- and three-body contributions \cite{Barker1971,Attard1992a}. Unlike those studies, which focus on equilibrium thermodynamic properties in large ensembles, the present work considers a minimal three-body system and investigates its real-time dynamics via a Lagrangian formulation. All quantities are nondimensionalized and therefore correspond to the literature values up to a rescaling by the characteristic LJ length and energy parameters.

\textit{Scope and physical validity:}
The AT expression arises under the assumption of negligible overlap between the atomic charge distributions. It is therefore an asymptotic long-range approximation and not a complete three-body potential. At sufficiently short separations, exchange and overlap effects become comparable to and eventually exceed the triple-dipole contribution \cite{Polymeropoulos1984}. Accordingly, the AT interaction is expected to provide a quantitatively reliable description only once the atoms are sufficiently separated that charge overlap is negligible. Hence for the isolated AT term, at short range the contribution is comparatively large but its underlying assumptions are violated, and at long range the assumptions hold but the term is a very small correction to the dominant pair interaction LJ.

Much of the configuration space we sample therefore lies outside the regime where AT quantitatively describes real rare-gas atoms, and we make no claim to such realism. Both the isolated AT system and the LJ$+$AT composite are physics-inspired analytic models, not meant to be realistic physical descriptions. They are used to test whether symmetry-encoded networks can learn genuinely coupled two- and three-body structure from dynamics. Our conclusions regarding learnability, representation, and identifiability therefore concern the learning problem itself, they do not concern the quantitative modeling of any specific material.

\section{Methods}
\label{sec:methods}

\begin{figure}[t]
    \centering
   \includegraphics[width=0.78\linewidth]{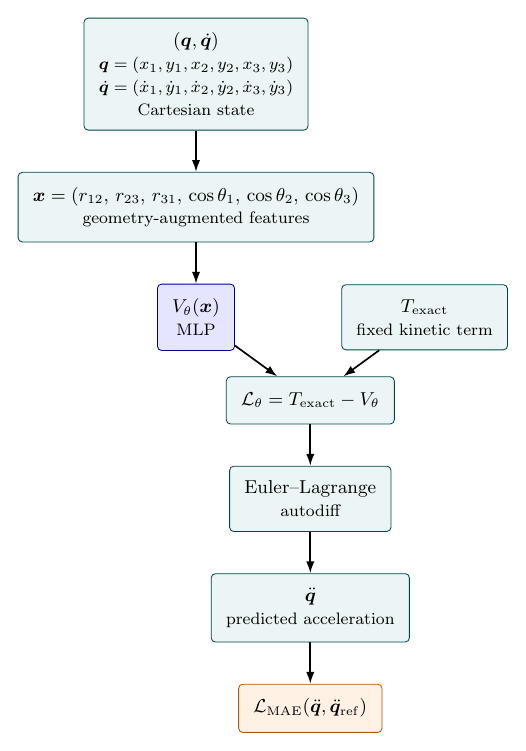}
    \caption{Schematic of the fixed-kinetic Lagrangian network: known structure (kinetic term, symmetries) is encoded exactly, and only the interaction potential is learned. The potential is not supervised directly, the training signal reaches it only through the accelerations it generates.}
    \label{fig:schematic}
\end{figure}

Our approach is summarized in Fig.~\ref{fig:schematic}: a partially parameterized Lagrangian in which the kinetic term is prescribed analytically and only the interaction potential is learned, as an unknown function of the three-body geometry. Dynamics follow by differentiating this Lagrangian through the Euler--Lagrange equations (Sec.~\ref{sec:lnn-background}), and training matches the resulting accelerations to reference data. The two structural choices, the fixed kinetic term and the augmented input reparameterization, are the subject of this section, else is mostly standard and stated briefly below.

\begin{table}[t]
\caption{Lagrangian parameterizations relevant to this work.}
\label{tab:variants}
\begin{ruledtabular}
\begin{tabular}{ll}
Method & Learned components \\
\colrule
DeLaN \cite{Lutter2019} & $M_\theta(\bm q)$ and $V_\theta(\bm q)$ \\
LNN \cite{Cranmer2020} & entire $\mathcal{L}_\theta(\bm q,\dot{\bm q})$ \\
Hamzao\u{g}ullar\i{} \& \"Ozak\i n \cite{hamza2026} &
entire $\mathcal{L}_\theta(\bm q,\dot{\bm q})$, \\
  & Hessian-regularized \\
This work & analytic $T(\dot{\bm q}) + V_\theta(\bm x(\bm q))$ \\
\end{tabular}
\end{ruledtabular}
\end{table}

$V_\theta$ is a multilayer perceptron with GELU activations as utilized by Hamzao\u{g}ullar\i{} \& \"Ozak\i n \cite{hamza2026}, and training minimizes the mean absolute error of predicted accelerations,
\begin{equation}
  \mathcal{L}_{\mathrm{train}}
  = \frac{1}{N}\sum_{i=1}^{N}
    \left\lvert \ddot{\bm q}_i^{\,\mathrm{pred}} - \ddot{\bm q}_i^{\,\mathrm{true}} \right\rvert.
  \label{eq:loss}
\end{equation}
Training and validation sets (120k and 24k configurations) sample pairwise distances from a log-uniform mixture of dense and sparse regions. The sampling distribution is designed to maximize exposure to the narrow regime where the interaction remains both informative and numerically stable, hence it's an adaptive allocation of data to where the learning signal lives, details are described in Sec.~\ref{sec:experiments}. All remaining details: sampling parameters and construction, optimizer and schedule are given in Appendices~\ref{app:dataset} and~\ref{app:training}; Table~\ref{tab:variants} situates our parameterization among existing Lagrangian networks.

\subsection{Fixed kinetic term}
\label{sec:fixed-kinetic}

We employ a partially parameterized Lagrangian in which only the unknown potential energy is represented by a neural network:
\begin{equation}
  \mathcal{L}_\theta
  = T_{\mathrm{exact}}
  - V_\theta(\bm x(\bm q)),
  \qquad
  T_{\mathrm{exact}}
  = \frac{m}{2}\sum_i \lvert\dot{\bm q}_i\rvert^2 ,
  \label{eq:fixedL}
\end{equation}
where $\bm x(\bm q)$ is described in Sec.~\ref{sec:input-rep}. Unlike systems whose generalized-coordinate formulation induces a configuration-dependent mass matrix $M(\bm q)$, our problem is formulated in Cartesian coordinates with known particle masses, so the kinetic energy takes a fixed quadratic form with a constant mass matrix $M=mI$. With $T$ prescribed, the mixed Hessian in Eq.~\eqref{eq:EL} vanishes and the Euler--Lagrange equations reduce identically to
\begin{equation}
  m\,\ddot{\bm q}
  =
  -\nabla_{\bm q}
  V_\theta\!\bigl(\bm x(\bm q)\bigr),
  \label{eq:reduction}
\end{equation}
so the learning problem becomes the recovery of a scalar function from its gradient field, well-posed up to an additive constant wherever the data cover configuration space.

With this decomposition, we omit the need to infer the known quadratic kinetic form, and concentrate the representational capacity of the neural network on the unknown interaction physics. Our architecture may therefore be regarded as a domain-specialized variant of a general Lagrangian neural network, in which a priori physical knowledge is incorporated analytically.

Within the class of systems satisfying the conditions described, this specialization removes the freedom that the true Lagrangian already did not have, so it loses nothing relevant relative to a fully parameterized Lagrangian network.

\subsection{Input reparameterization}
\label{sec:input-rep}

The potential network receives the augmented coordinates
\begin{equation}
  \bm x = \left(r_{12},\, r_{23},\, r_{31},\, \cos\theta_1,\, \cos\theta_2,\, \cos\theta_3\right),
  \label{eq:features}
\end{equation}
computed internally from the Cartesian positions. Since the first three coordinates alone are sufficient to uniquely specify any planar three-body geometry intrinsically, the angular coordinates introduce no additional geometric information, as the interior angles are uniquely determined by the three pairwise distances through the law of cosines.  We nevertheless include both distances and angles because this mixed representation exposes radial and angular structure explicitly, an inductive bias that has been used in machine-learned interatomic potentials (Sec.~\ref{sec:related-mlp}) and is supported empirically.

Among the possible angular features, $\cos\theta$ is the natural choice because it is one-to-one on the physical range $\theta\in[0,\pi]$. With this choice, no input normalization is applied, since over the sampled ranges the internal coordinates are already $O(1)$.

Trivially, the augmented coordinates carry no information absent from the Cartesian positions. The mapping is a symmetry-adapted reparameterization of the same configuration. From the network's perspective, the interaction remains an unknown function inferred from motion data. The empirical evidence in Sec.~\ref{sec:exp_at} supports this interpretation: the same representation recovers a purely radial potential when trained on Lennard--Jones data (and seperately on Newtonian data), without introducing spurious angular dependence, while the identical featurization also succeeds on the composite LJ$+$AT system. The representation therefore acts as a generic description of three-body geometry and a generic inductive bias for three-body potentials.

Finally, velocities never enter $V_\theta$: they appear only in the fixed kinetic term and in the Euler--Lagrange algebra of Eq.~\eqref{eq:EL}. Both the learned and the reference accelerations are obtained by differentiating the respective potentials with respect to the Cartesian positions, automatic differentiation passes through the featurization of Eq.~\eqref{eq:features}, so the network is evaluated on internal coordinates while the dynamics remain Cartesian.

\section{Experiments}
\label{sec:experiments}

We divide the experiments into two groups. The first isolates the three-body interaction entirely, where the training data are generated by the AT term alone, hence the question of whether the network can represent purely non-additive structure is posed in its cleanest form, free of any pairwise background (Sec.~\ref{sec:exp_at}). The second restores the composite setting of practical interest, a dominant pairwise LJ interaction with three-body corrections of varying relative strength, which shifts the question from representability to identifiability (Sec.~\ref{sec:exp_ljat}).

Within the isolated study we evaluate at three stages: (i) potential landscapes and force fields, the instantaneous field-level comparison between learned and analytic dynamics; (ii) the functional geometric dependences, radial and angular, that define the AT interaction in closed form; and (iii) trajectory rollouts, in which model errors accumulate under time integration. The stages are aimed to be complementary, hence each provides information from a different perspective.  Reporting all three is aimed to disambiguate the failure modes correspondingly.

\begin{figure}[t]
    \centering
    \includegraphics[width=\linewidth]{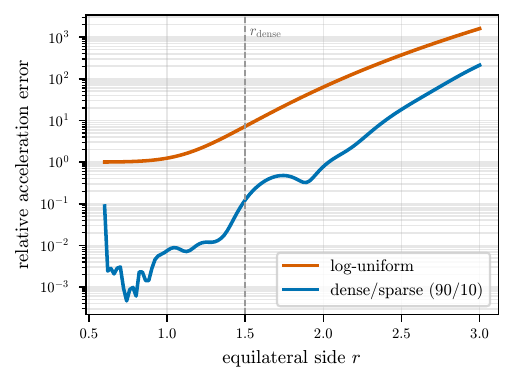}
    \caption{Relative acceleration error along an equilateral configuration sweep for models trained with the proposed dense/sparse sampler and with plain log-uniform sampling. The dense/sparse model reaches relative errors of $8.8\%$, $0.8\%$, and $13.5\%$ at $r=0.6$, $1.0$, and $1.5$, against $100.4\%$, $124.9\%$, and $747.2\%$ for the log-uniform model over the same points. Both models degrade further into the tail, but the gap persists, an order-of-magnitude advantage holds even where both models are far from quantitatively accurate.}
\label{fig:sampling_ablation}
\end{figure}

\textit{Dataset Design:} Because $|\nabla V|$ spans several orders of magnitude (Sec.~\ref{sec:theory}), the
loss in Eq.~\eqref{eq:loss} is dominated by compact, small-$r$ configurations, where
the AT acceleration grows as $r^{-10}$. This regime carries the information that
fixes the interaction and governs rollout accuracy, so the sampler must visit it
often. Log-uniform sampling of separations does not, only $28\%$ of its configurations have at least two separations below
$r_{\rm dense}=1.5$, compared with $84\%$ for the dense/sparse sampler. We note that the
dense/sparse sampler still keeps mixed and extended geometries, since trajectories
visit them too. To test the effect, we trained two models identical in every respect
except the sampler (Appendix~\ref{app:dataset}). Along an equilateral sweep, the
dense/sparse model reproduces the analytic acceleration to better than $1\%$ for
$0.62\lesssim r\lesssim1.2$ and to within $\approx\!14\%$ over the whole dense range $r\le1.5$, whereas the log-uniform model has a relative
error of at least $\approx\!100\%$ over the whole sweep
(Fig.~\ref{fig:sampling_ablation}). Covering the same range of $r$ is
therefore not enough, and we claim that what matters is how often the strong-interaction regime is
visited. This is also directly affecting the balance of the loss, since compact configurations' errors dominate the loss even when they are too rare to be fitted, as under log-uniform sampling, and the model ends up fitting neither regime. The cost is accuracy at large $r$, where the force is already small and dynamically
irrelevant. Both models expectedly degrade there, but the dense/sparse model remains roughly an
order of magnitude more accurate,  and the $10\%$ of separations drawn from the sparse
range keep that region from being left entirely unlearned.

\subsection{Isolated three-body interaction}
\label{sec:exp_at}

Unless stated otherwise, results refer to the reference pure-AT model ($\nu=1.0$, $r\in[0.6,5.0]$; Appendix~\ref{app:dataset}), evaluated against the analytic dynamics. 

\subsubsection{Potential landscapes and force fields}
\label{sec:exp_at_fields}

\begin{figure*}[!t]
    \centering
    \includegraphics[width=0.95\linewidth]{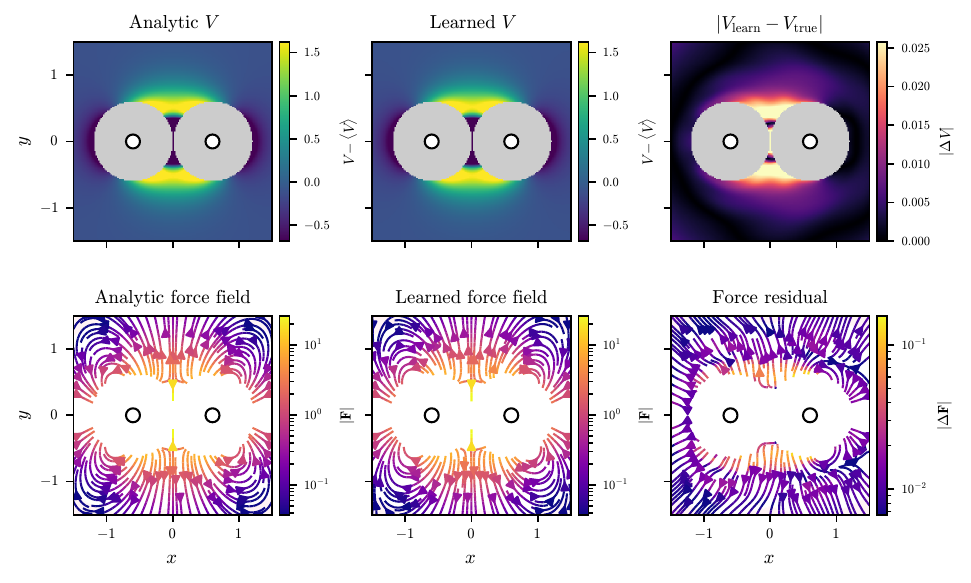}
    \caption{Field-level comparison for the isolated AT interaction. Two particles (white markers) are held fixed while the third is evaluated over a two-dimensional grid of positions. Top row: analytic potential, learned potential, and their absolute difference. The potentials are centered by subtracting their spatial mean, reflecting the arbitrary additive constant in the potential energy. Bottom row: corresponding acceleration fields shown as streamlines colored by magnitude, together with the pointwise acceleration residual. The region where any pairwise distance falls below the training cutoff $r_{\min}$ is masked. The pink overlay  indicates regions that were sampled less densely during training. The two fixed particles are separated by $r_{12}=1.20$; axes are in reduced units.}
    \label{fig:fields}
\end{figure*}

To assess the learned interaction, we first compare the complete potential landscape and corresponding acceleration field over a two-dimensional slice of configuration space. Two particles are held fixed while the third is evaluated on a uniform grid of positions, producing a continuum of three-body configurations (Fig.~\ref{fig:fields}). This simultaneously inspects the local geometry of the potential surface and the vector field generated by its gradient.

Figure~\ref{fig:fields} shows close agreement between the learned and analytic potentials throughout the sampled domain, with the residuals remaining small relative to the overall variation of the landscape. The learned acceleration field also reproduces the structure of the analytic field, including the locations of attractive and repulsive flow regions. Aggregated over the displayed domain (which is mostly but not entirely in the dense interval), the relative L1 error is $1.3\%$ in the potential and $1.1\%$ in the acceleration magnitude (magnitude-weighted, App.~\ref{app:eval_metrics}).

The potentials are centered before comparison because the zero of potential energy is arbitrary: only potential differences, equivalently their gradients and the resulting accelerations, are physically meaningful. The largest residuals in the figure appear, in absolute terms, right outside the excluded region near $r_{\min}$, but this is simply because the potential itself is largest there. Measured relatively, this near-field band is in fact the most accurate part of the domain ($\approx0.9$--$1.0\%$ for $r\in[0.6,1.2]$). 

Since the configurations on this evaluation also include many different but not parameterized variations of angular and radial arrangements, it provides an insightful global picture of the learned interaction, which the next two subsections' one dimensional analyses complement.

\subsubsection{Radial dependence}
\label{sec:exp_at_radial}

\begin{figure*}[t]
    \centering
    \includegraphics[width=0.85\linewidth]{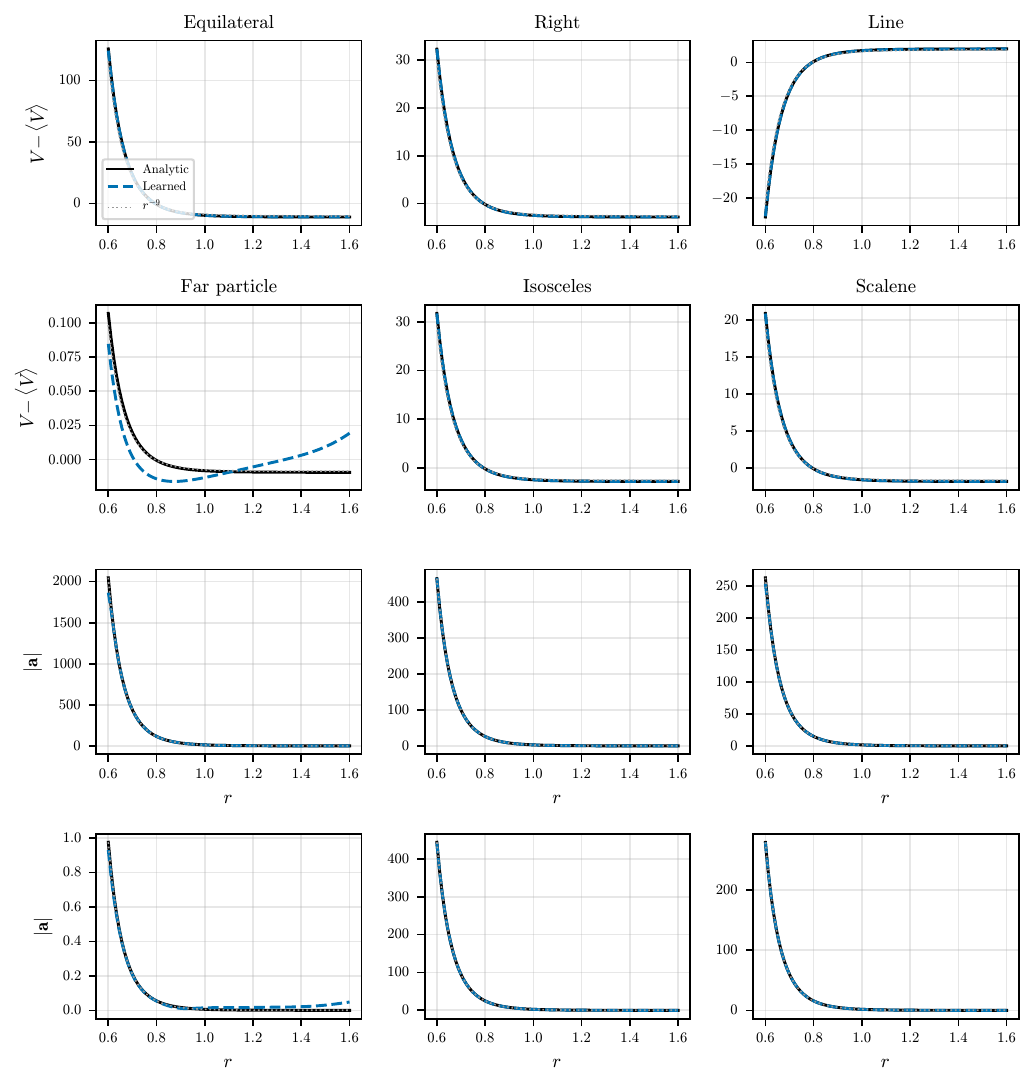}
    \caption{Radial dependence of the learned Axilrod--Teller interaction. Each column corresponds to a fixed reference triangle, with all remaining configurations obtained by uniformly scaling that geometry. The horizontal axis $r$ is the overall scale factor multiplying a fixed unit shape, so every pairwise distance grows $\propto r$. For the equilateral family $r$ is the side length, for the others the linear size of the reference triangle. Because rescaling leaves all angles fixed, each family must follow $V\propto r^{-9}$, $|\mathbf a|\propto r^{-10}$. The far-particle family is the sole visibly separating curve. For this family two of the three sides lie in the sparsely sampled range $r>r_{\rm dense}$ (App.~\ref{app:dataset}), where the net three-body force is smallest, so a small absolute residual becomes a large relative one. Rows 1--2: learned and analytic potentials, rows 3--4: corresponding acceleration magnitudes.}
    \label{fig:radial}
\end{figure*}

Uniformly scaling a fixed reference triangle preserves its shape while varying only its overall size. We evaluate six representative reference geometries, comparing learned and analytic potential and
acceleration magnitude at each scale by dividing the potential and acceleration values to the fixed-by-geometry truth of the angular term. As shown in Fig.~\ref{fig:radial}, agreement is close for the five families, relative L1 error (magnitude-weighted over the sweep) of $0.25\%$ in potential and $0.66\%$ in acceleration, averaged across Equilateral, Right, Line, Isosceles, and Scalene. The far-particle family is the exception, with relative L1 error of $87.6\%$ (potential) and $21.6\%$ (acceleration). As the third atom recedes, the net three-body contribution falls fastest for this geometry, hence a small absolute residual becomes a large relative one, consistent with the visible separation in Fig.~\ref{fig:radial} and not a
true functional-form discrepancy. 
\begin{figure}[t]
    \centering
    \includegraphics[width=0.95\linewidth]{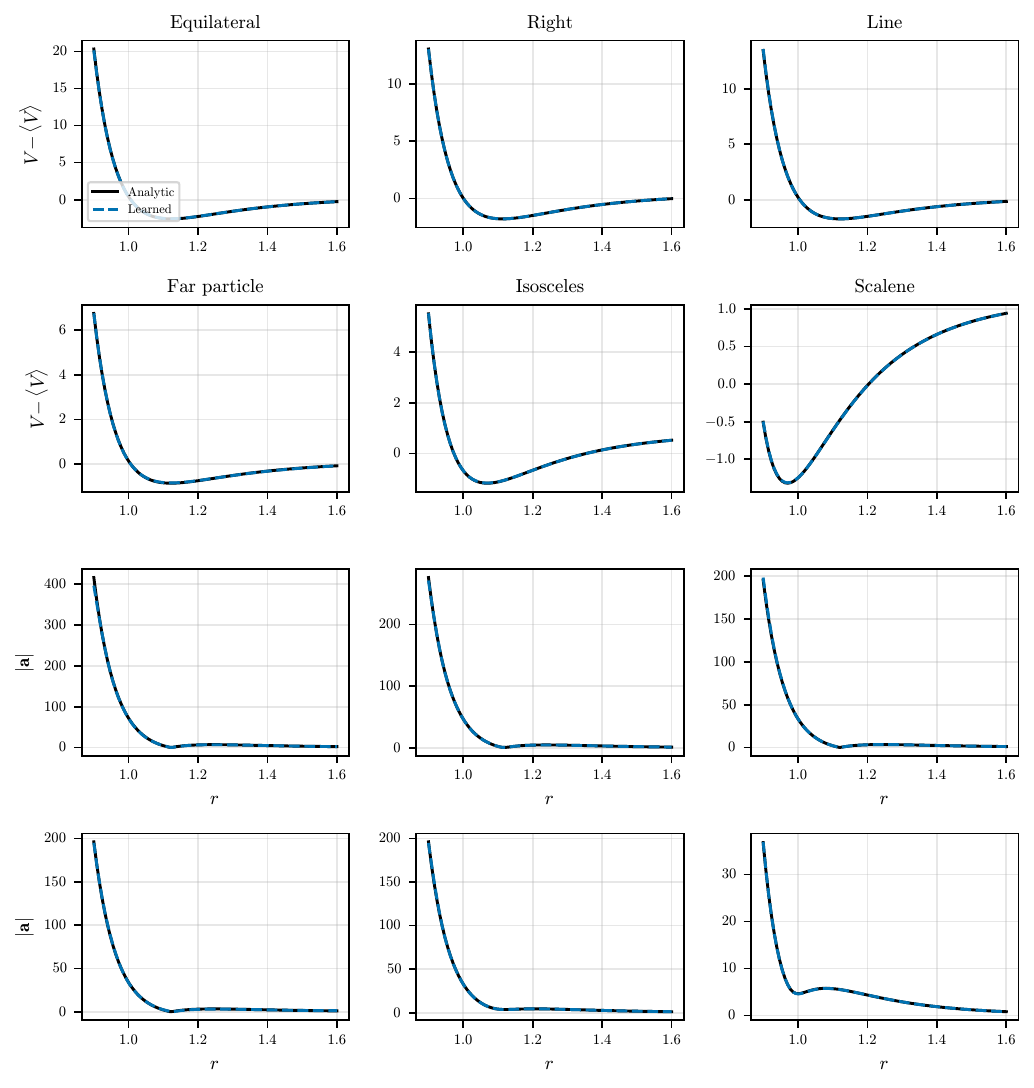}
    \caption{Pure Lennard--Jones control. The same geometric input representation, including both pairwise distances and angle cosines, is trained on a purely pairwise interaction. The learned potential remains in good agreement with the analytic LJ radial dependence across all reference geometries, indicating that the additional angular inputs do not introduce spurious shape dependence. Rows 1--2: learned and analytic potentials, rows 3--4: corresponding acceleration magnitudes.}
    \label{fig:lj_control}
\end{figure}

\paragraph*{Pure LJ control.} 
A natural concern is that including the angular coordinates $\cos\theta_i$ among the inputs might bias the network toward learning an artificial shape dependence, even when the underlying interaction is purely pairwise. Figure~\ref{fig:lj_control} addresses this directly. Using the identical architecture, same geometric representation, and training protocol, but training on a purely Lennard--Jones system, the learned potential remains in good agreement with the analytic radial dependence with a relative L1 error of $0.45\%$ (potential) and $0.76\%$ (acceleration) across all six reference geometries. The representation therefore acts as a neutral symmetry-adapted parameterization and an inductive bias toward many-body angular structure.

\begin{figure}[t]
    \centering
    \includegraphics[width=0.72\linewidth]{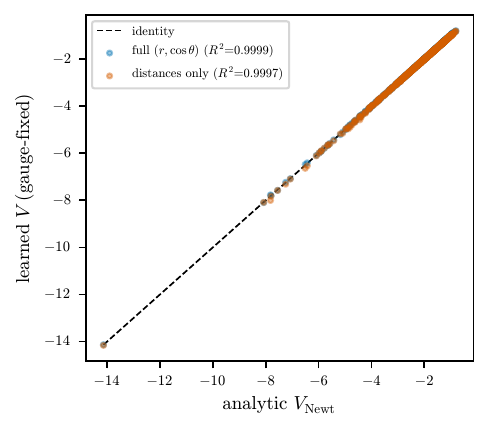}
    \caption{Newtonian gravity control. Learned versus analytic potential
    (after gauge-fixing by a single additive offset) over a set of random three-body
    configurations within the training range, for the full $(r,\cos\theta)$ model and the distances-only
    ablation. Both recover the pairwise $1/r$ potential essentially exactly
    ($R^2 = 0.9999$ and $0.9997$), with mean relative acceleration errors of
    $2.6\%$ and $2.5\%$, confirming that the geometric featurization fits a
    radial law nothing like the Axilrod--Teller term's, and that the angular
    inputs stay inert when the interaction carries no angular signal.}
    \label{fig:newtonian_control}
\end{figure}

\paragraph*{Newtonian control.}
The same conclusion holds for a structurally very different pairwise
interaction. Trained on data from three particles under Newtonian gravity
($V = -G\sum_{a<b} m_a m_b / r_{ab}$, which is a $1/r$ potential with no three-body
term), the identical $(r,\cos\theta)$ architecture recovers the gravitational
force law to a mean relative acceleration error of $2.6\%$ and reproduces the
potential up to its additive gauge constant essentially exactly
($R^2 = 0.9999$, App.~\ref{app:eval_metrics}, Fig.~\ref{fig:newtonian_control}). Gravity is a demanding second control because its radial law is not similar to the AT term's. Where AT falls as $r^{-9}$ and LJ as a steep $12$--$6$ wall, gravity is long-ranged and shallower, yet the same featurization fits it without modification. A distances-only ablation, with three radial inputs, no angle
cosines, trained on the identical data performs indistinguishably ($2.5\%$ mean relative acceleration error, $R^2 = 0.9997$), confirming that the angular inputs are simply not needed when the interaction carries no angular signal, and that their presence in the full model neither helps nor hurts a purely radial target.

\subsubsection{Angular dependence}
\label{sec:exp_at_angular}

\begin{figure}[t]
    \centering
    \includegraphics[width=\linewidth]{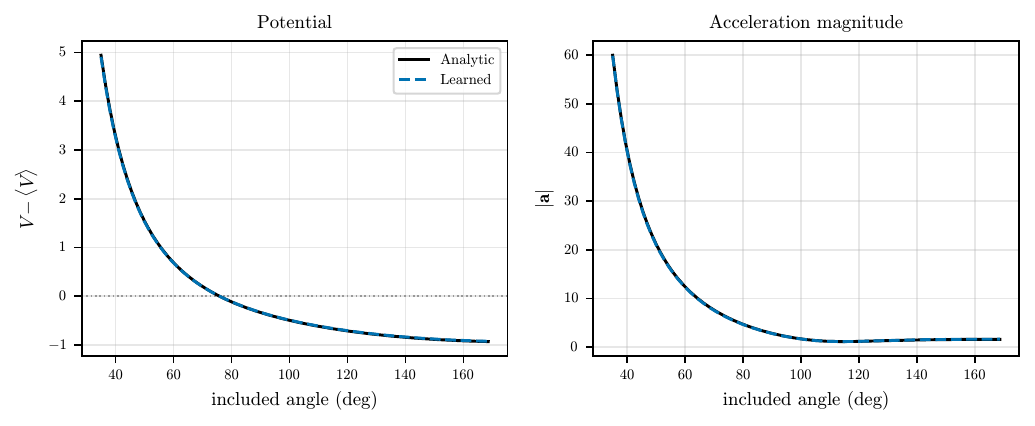}
    \caption{Angular dependence of the learned interaction at fixed length scale. Particle 1 is at the origin, particle 2 at $(s,0)$, and particle 3 at $s(\cos\theta,\sin\theta)$, so the two sides meeting at particle 1 are held equal, $r_{12}=r_{13}=s$, while the swept apex angle $\theta$ sets the third side $r_{23}=2s\sin(\theta/2)$ and hence the shape. Thus $s$ fixes the characteristic radial scale while $\theta$ alone varies the triangle shape, isolating the angular factor $\Phi$ at (near-)constant size. (Left) Potential, after subtracting its mean. (Right) Corresponding acceleration magnitude.}
    \label{fig:angular}
\end{figure}

The more characteristic signature of the AT is the angular modulation encoded by the factor $\Phi$. To isolate this dependence in an interpretable way, we fix the characteristic length scale and vary only the triangle shape by continuously changing the included angle while keeping one side of the reference triangle fixed (Fig.~\ref{fig:angular}). As the angle is swept, the learned potential and acceleration magnitude remain in good agreement of mean relative errors $0.81\%$ in potential and $1.06\%$ in acceleration, with their analytic counterparts throughout the angular range, including the transition between the repulsive and attractive branches. 

\subsubsection{Trajectory rollouts}
\label{sec:exp_at_traj}
\begin{figure*}[t]
    \centering
    \begin{subfigure}{\textwidth}
        \centering
        \includegraphics[width=0.8\textwidth]{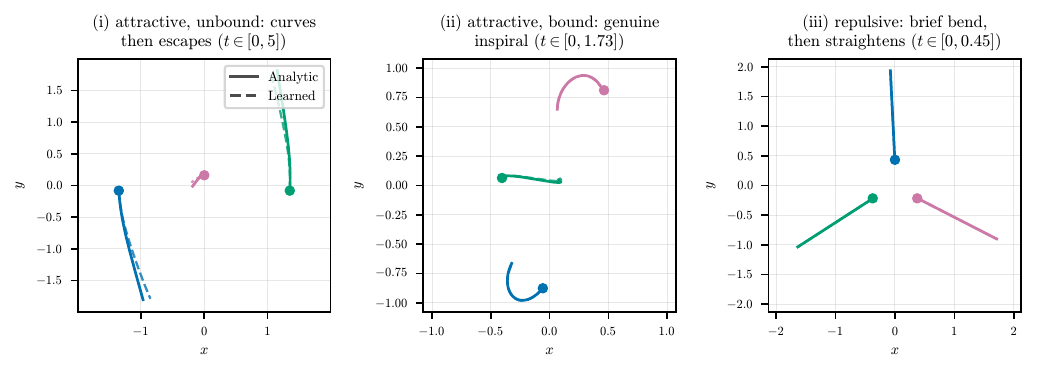}
        \caption{Three individual rollouts, each showing a qualitatively different
        outcome (reference solid, learned dashed, particles colour-coded, dots mark
        initial positions): (i) an attractive ($\Phi<0$) triangle with enough kinetic
        energy to remain unbound, which curves before eventually separating,
        (ii) an attractive triangle with too little energy to escape, spiraling in
        to a direct collision, (iii) a repulsive ($\Phi>0$) triangle, shown over a
        short window to isolate its brief initial bend before the trajectory
        straightens into free separation.}
        \label{fig:traj_a}
    \end{subfigure}

    \vspace{4pt}

    \begin{subfigure}{\textwidth}
        \centering
        \includegraphics[width=0.6\textwidth]{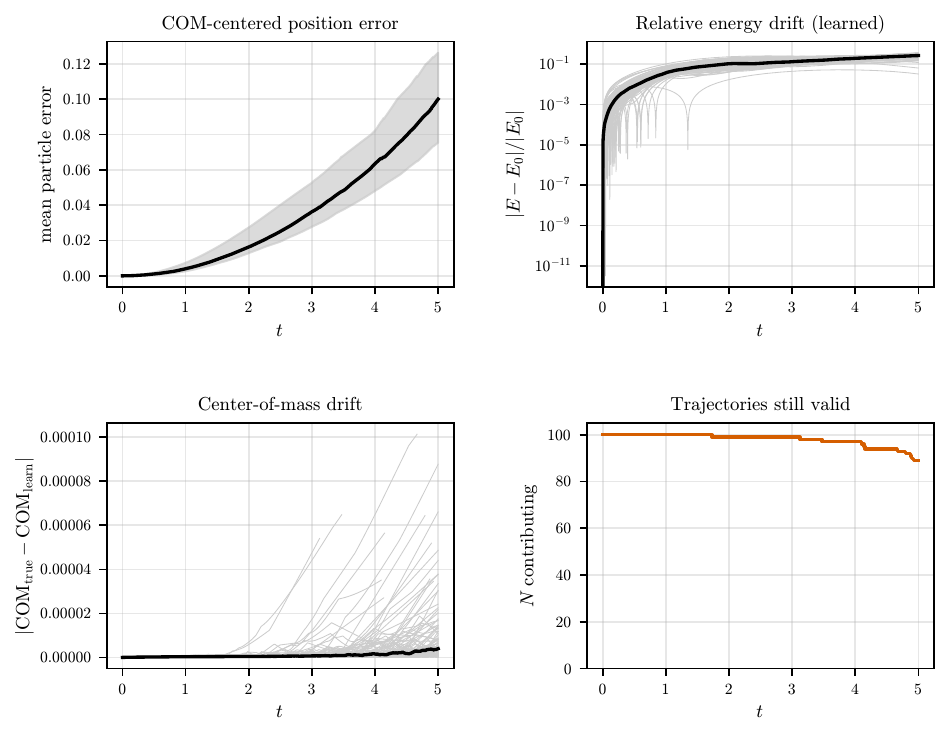}
        \caption{Ensemble statistics over $N=100$ random initial conditions, each
        trajectory contributes to a given time only until it leaves the
        training-valid region. Top left: center-of-mass-corrected position error (median, 10--90\% band shaded). Top right: relative energy error $\varepsilon_E$ of Eq.~\eqref{eq:energy_error} (individual trajectories in grey, median in black, log scale). Bottom left: drift between the reference and learned centers of mass, as a sanity check. Bottom right: number of trajectories still inside the training domain.}
        \label{fig:traj_b}
    \end{subfigure}

    \caption{Trajectory-level evaluation of the learned dynamics.}
    \label{fig:traj}
\end{figure*}

We integrate the learned Euler--Lagrange equations from initial conditions drawn at
random within the training bounds (unless noted otherwise) and compare the resulting
trajectories against reference trajectories obtained by integrating the exact
analytic interaction (Fig.~\ref{fig:traj}). 

Fig.~\ref{fig:traj}(a) shows three
qualitative outcomes side by side: an unbound attractive case (i), a bound attractive
case that collapses (ii), and a repulsive case whose curvature is confined
to a brief initial window before it straightens into free separation (iii).

Over the $N=100$ ensemble, $89$ of which remain inside the training domain for
the entire window, the rest contributing until the moment they leave it, and we
report three statistics. Position error is computed after subtracting the
center-of-mass displacement of each frame, so that rigid translation, to which the
interaction is blind by construction, is not conflated with actual interaction
error. Its ensemble median stays at $1.9\%$ of the mean initial particle separation, rising to $7.1\%$ by the time the window closes. Because the
learned potential depends on positions only through translation- and
rotation-invariant features, the total force on the system sums to zero identically
and momentum is conserved up to integration error alone. The center-of-mass drift
panel confirms this ($4.8\times10^{-7}$ over the whole window, rising to
only $5.2\times10^{-6}$ by its end), which verifies that the position error above is not
concealing a hidden momentum failure. 

Finally, we define the \emph{relative energy error} as the true (analytic) energy functional
evaluated on the learned rollout, compared against the same functional evaluated on the
ground-truth rollout at the same instant, normalized by the ground truth's initial energy:
\begin{equation}
    \varepsilon_E(t) \;=\; \frac{\bigl|\,E_{\mathrm{true}}\bigl(x_{\mathrm{learn}}(t)\bigr)
    - E_{\mathrm{true}}\bigl(x_{\mathrm{true}}(t)\bigr)\,\bigr|}
    {\bigl|\,E_{\mathrm{true}}\bigl(x_{\mathrm{true}}(0)\bigr)\,\bigr|} .
    \label{eq:energy_error}
\end{equation}
Both rollouts are integrated from the same initial state, so $\varepsilon_E(0)=0$ by construction. What Eq.~\eqref{eq:energy_error} tracks from there is how far the two trajectories' energies pull
apart as the rollout proceeds. Since the exact dynamics conserve energy, $E_{\mathrm{true}}\bigl(x_{\mathrm{true}}(t)\bigr)
= E_{\mathrm{true}}\bigl(x_{\mathrm{true}}(0)\bigr) \equiv E_0$ , so Eq.~\eqref{eq:energy_error} equals $|E-E_0|/|E_0|$ with
$E = E_{\mathrm{true}}\bigl(x_{\mathrm{learn}}(t)\bigr)$, the form plotted in Fig.~\ref{fig:traj_b}. 

Over the ensemble, $\varepsilon_E$ has a median of $11.5\%$ across the
whole window, rising to $26.6\%$ by its end.

This is the most demanding of our three measures, because it is sensitive to the trajectory's full
history through a potential whose local steepness varies over many orders of magnitude. As the
learned and true rollouts separate in configuration space, by exactly the position and velocity
error, $E_{\mathrm{true}}$ is evaluated at two increasingly different
states, and the same, modest, bounded separation is weighted by the local force at each instant.
Because that force varies by orders of magnitude across the accessible configuration space, an
identical positional error is amplified very differently depending on where it occurs. This is also why the model tracks the system's \emph{shape} far
more tightly than it appears to conserve \emph{energy} over a long rollout.

We note that Eq.~\eqref{eq:energy_error} is a statement about \emph{fidelity to the true dynamics},
not about the model's own internal consistency. The learned Lagrangian is time-independent, so it conserves \emph{its own} energy exactly, up to integrator error evaluated with
the network's own learned potential, the learned rollout's energy is
constant to within integrator precision ($\sim 10^{-6}$ relative error). The large values of
$\varepsilon_E$ therefore do not indicate a violation of the model's structural conservation law,
they indicate that the potential it conserves is a close, but imperfect, approximation of the true
one, and that this small approximation error compounds, through the steepness described above, into
a large discrepancy against the true energy over a long rollout.

\subsubsection{The canonical configurations of Axilrod and Teller}
\label{sec:exp_at_special}
\begin{figure*}[t]
    \centering
    \includegraphics[width=0.8\textwidth]{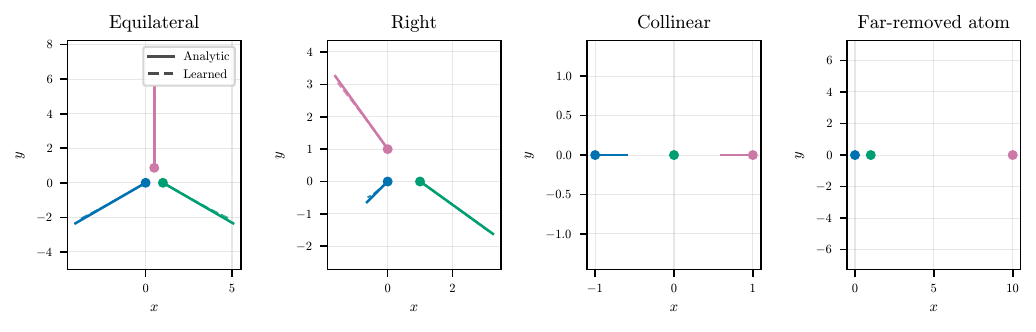}
    \caption{Reference (solid) and learned (dashed) dynamics initialized from the four triangle geometries used by Axilrod and Teller \cite{Axilrod1943} to illustrate the sign of $\Phi$: equilateral, right-angled, collinear, and the far-removed-atom limit. Trajectories are shown until a pair separation falls below the training cutoff $r_{\min}$.}
    \label{fig:special}
\end{figure*}

As a final, gestural check, we revisit the four triangle geometries Axilrod and Teller used to illustrate the shape-dependence of $\Phi$'s sign \cite{Axilrod1943}: the equilateral triangle ($\Phi=11/8$, maximally repulsive), the right triangle ($\Phi=1$, repulsive), the collinear configuration ($\Phi=-2$, attractive), and the far-removed-atom limit, where the sign is set by a single angle through $1-3\cos^2\theta_1$. The learned dynamics reproduce the repulsive break-up of the compact triangles and the attractive collapse of the collinear one (Fig.~\ref{fig:special}). The far-removed configuration lies outside the training range ($r>r_{\max}$), where the learned interaction is not actually constrained.

\subsection{Three-body corrections on a pairwise background}
\label{sec:exp_ljat}

Having shown that the isolated three-body interaction is representable
(Sec.~\ref{sec:exp_at}), we restore the pairwise Lennard--Jones background and ask
the question that matters for realistic simulation: when does a three-body correction
leave a signature in the dynamics that a learned model can actually resolve? We pose the answer as governed by a single comparison, the size of the AT contribution against
the model's own fitting-error floor. We test it from both sides, by varying
the signal (two couplings) and by removing the floor (a control in which the pairwise
part is supplied exactly).

\begin{figure*}[t]
    \centering
    \includegraphics[width=0.8\textwidth]{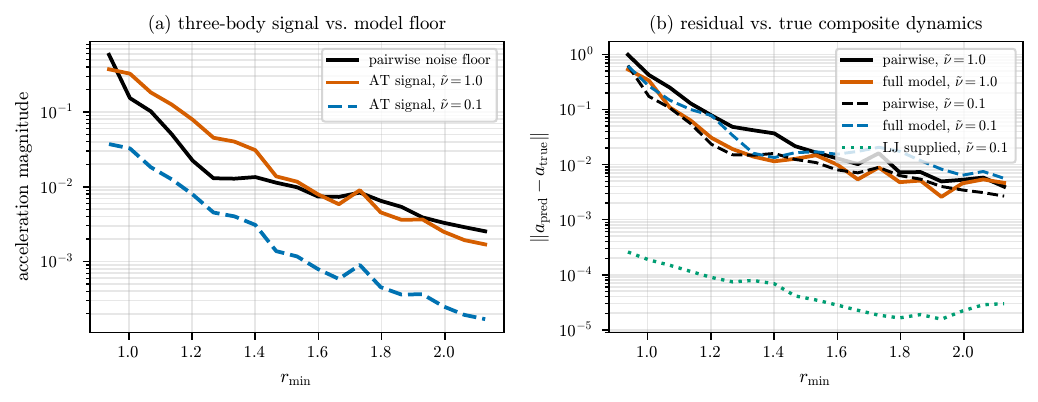}
    \caption{Identifiability of the three-body term on a pairwise background, versus
    the inner separation $r_{\min}$. All curves are binned medians of an acceleration
    magnitude $|a|$ over validation configurations. (a) The AT contribution the pairwise
    model omits, $|a_\mathrm{AT}|=\tilde\nu\,|a^{(1)}_\mathrm{AT}|$, at the physical
    ($\tilde\nu=0.1$) and amplified ($\tilde\nu=1.0$) couplings, against the pairwise
    (LJ-only) model's error floor. (b) Residual $|a_\mathrm{pred}-a_\mathrm{true}|$
    against the true composite dynamics for the pairwise-only and full learned-potential
    models at both couplings, and for the control at $\tilde\nu=0.1$ in which the LJ
    term is supplied analytically and only the three-body part is learned.}
    \label{fig:ljat}
\end{figure*}

Throughout this section we use one metric: the acceleration error
$|a_\mathrm{pred}-a_\mathrm{true}|$ of a configuration, in reduced units, summarized by
its median over the argon-reference validation configurations ($r_{\min}\!\ge\!0.9$).
The median is robust to the steep repulsive wall, where a few close configurations
carry very large accelerations. Because the acceleration is linear in the AT coupling
and the LJ part is common to both systems, the exact contribution a pairwise model
omits is $|a_\mathrm{AT}|=\tilde\nu\,|a^{(1)}_\mathrm{AT}|$, measured in the same units.
We compare it with the pairwise model's \emph{error floor}: the LJ-only model's
residual against the pairwise dynamics it was trained on, i.e.\ its irreducible
fitting error at our architecture and data budget. Its median is $0.045$
(Fig.~\ref{fig:ljat}a).

\subsubsection{Physical coupling}

At the physical $\tilde\nu\approx0.1$ of argon \cite{Attard1992a,Attard1992b} the
three-body term is not identifiable by a model that learns the full potential. Its
median magnitude is $0.009$, five times below the pairwise floor, and it lies below
the floor in $85\%$ of configurations: the signal sits inside the error of the
pairwise fit. Accordingly, the full model does not improve on the pairwise one: its
median residual against the true composite dynamics ($0.091$) is larger than the
pairwise model's ($0.053$; Fig.~\ref{fig:ljat}b).

The obstruction is the pairwise floor, not the three-body term. To show this, we
train the same architecture on the same data with the LJ term supplied analytically,
so that only the three-body part is learned,
$V = V_\mathrm{LJ} + \tilde\nu\, g_\theta(r_{ij},\cos\theta_i)$.\footnote{The fixed
prefactor $\tilde\nu$ only scales the network output and could be absorbed into its
last layer. A pure-AT model at the same coupling, trained without it, reaches a median
residual of $5.4\times10^{-4}$ on the same configurations, $16\times$ below the
three-body signal.}
Without the pairwise fitting error, the three-body term is learned: the median
residual is $1.1\times10^{-4}$, $80\times$ below the signal and $400\times$ below the
pairwise floor, and lies below both in every configuration (Fig.~\ref{fig:ljat}b).
The null result therefore reflects the size of the pairwise fitting error relative to
the three-body term, and lowering that error, for instance with an exact or pretrained
pairwise potential or a larger model, moves the identifiability boundary toward the
physical coupling.

\subsubsection{Amplified coupling}

At $\tilde\nu=1.0$ the comparison reverses, and the full model succeeds with no
decomposition supplied: a single learned potential, with no analytic LJ term and no
separation into pairwise and three-body parts, represents the complete interaction.
The AT contribution (median $0.087$) now stands above the pairwise floor in two-thirds
of configurations. The pairwise model, which cannot represent it, is left with a
median residual of $0.162$, while the full model brings it down to $0.065$, a
$2.5\times$ reduction that holds in $74\%$ of configurations and lands close to the
$0.045$ floor that limits even purely pairwise fitting (Fig.~\ref{fig:ljat}b). The learned total potential
thus captures the three-body correction to the resolution this architecture affords,
and recovers the pairwise and three-body orders together from the composite dynamics
alone.

Taken together, the three experiments bracket the identifiability boundary. Varying the coupling places the boundary between
$\tilde\nu=0.1$ and $1.0$. Removing the pairwise floor shows that the boundary is set
by that floor, and not by the representability of the three-body term, which this
architecture learns in isolation, as a residual on a known pairwise background, and,
once it clears the floor, inside a single total potential.

\section{Symbolic recovery of the interaction law}
\label{sec:exp_at_symbolic}

The preceding evaluations inspects the agreement of $V_\theta$ with $\AT$ pointwise,
radially, angularly, and dynamically, but $V_\theta$ itself remains an opaque
network. As a final, independent check, we ask whether a closed-form expression
can be extracted from $V_\theta$ directly, without assuming its functional form in
advance. We apply symbolic regression, where we chose PySR \cite{Cranmer2023PySR}, an open source symbolic regression library which uses a genetic evolutionary algorithm to fit symbolic expressions to data, because it is easily customizable. It searches
over algebraic expressions built from a fixed operator set for one that best
reproduces a target dataset, to $V_\theta$ evaluated on the geometric feature
vector $\bm x$ of Eq.~\eqref{eq:features}, with particles held at rest so that
$V_\theta(\bm x) = -\mathcal{L}_\theta(\bm x,\bm 0)$. Because $|V_\theta|$ spans
several orders of magnitude over the sampled domain (Sec.~\ref{sec:experiments}),
we use a relative squared error, stabilized in the standard way by adding a constant
offset $\eta=0.3$ to the target magnitude, so that the loss remains finite where
$V$ passes through zero:
\begin{equation}
    \mathcal{L}_{\mathrm{SR}}
    = \frac{1}{n}\sum_{i=1}^{n}
      \left(\frac{V_{\mathrm{SR},i} - V_{\theta,i}}{\lvert V_{\theta,i}\rvert + \eta}\right)^{2},
    \qquad \eta = 0.3
    \label{eq:sr-loss}
\end{equation}

Under a more standard squared error, the search is dominated by the small-$r$ core and
returns a degenerate fit, whereas the relative objective weights the full domain
and recovers structured expressions.

As a sanity check, applied to the \emph{analytic} $\AT$ under otherwise
identical settings, the search recovers Eq.~\eqref{eq:AT} exactly (coefficients
within $10^{-4}$ of the true values), confirming that the feature set, operator
set, and objective are together sufficient to find the true law before any claim
is made about $V_\theta$. 

Applied to $V_\theta$ itself, the search returns,
\begin{align}
    V_{\mathrm{SR}}(\bm x)
    &= c_0
    + \nu_{\mathrm{SR}}\,
      \frac{1.0 + 3.0\cos\theta_1\cos\theta_2\cos\theta_3}
           {(r_{12}r_{23}r_{31})^{3}}
    \nonumber\\
    &\quad + \delta_{\mathrm{SR}}\,(r_{23}+r_{31}),
    \label{eq:sr-recovered}
\end{align}
with the coefficients $\nu_{\mathrm{SR}}\approx1.0$, $c_0\approx0.9$, and $\delta_{\mathrm{SR}}\approx0.007$. Eq.~\eqref{eq:sr-recovered} reproduces
$\Phi$'s angular numerator and $r^{-9}$ radial law with the correct relative weight between them, matching
$V_\theta$ itself with $R^2\approx1.0$. The constant $c_0$ is not physical, it is the additive
gauge freedom already noted in Eq.~\eqref{eq:reduction}, absorbed by the search since it never
sees forces directly. Against the analytic $\AT$, the leading two terms of
Eq.~\eqref{eq:sr-recovered} achieve $R^2\approx0.98$, consistent with the small residual
discrepancy already visible in Secs.~\ref{sec:exp_at_fields}--\ref{sec:exp_at_angular}. The last term breaks the exchange symmetry that $\AT$ possesses exactly, consistent
with $V_\theta$ being only approximately symmetric under particle relabeling. Its
effect is negligible, dropping it leaves the fit to $V_\theta$ unchanged to five
decimal places in $R^2$, and its typical size is about $1\%$ of $|V_\theta|$ over
the fitting domain.


\paragraph*{Feature necessity.} Repeating the search with only the three
distances $(r_{12},r_{23},r_{31})$, withholding the angle cosines, the recovered
expression plateaus at $R^2\approx0.4$ against the analytic potential and
recovers only an approximate radial scaling, with no angular term. The search
cannot reconstruct $\Phi$ from distances alone within the operator budget used
here, even though the two representations carry the same information
(Sec.~\ref{sec:input-rep}). Augmenting the input with the cosines recovers $\AT$ exactly
under the same budget. This is independent evidence, from a different method
entirely, that the angular coordinates in Eq.~\eqref{eq:features} provide essential representational structure and are not a redundant convenience.

Robustness
of this recovery to the search's operator set, and full derivation details are given in
Appendix~\ref{app:symbolic}.

\section{Discussion and Conclusions}
\label{sec:discussion}

We have shown that Lagrangian Neural Networks, combined with physically informed sampling and domain-specific structure, can recover an irreducible three-body interaction from motion data alone. For the AT interaction, the recovery succeeds at several complementary levels.

The composite Lennard--Jones$+$AT system further illustrates the conditions under which such a higher-order interaction becomes dynamically identifiable. We recover the three-body contribution when it is sufficiently large relative to the residual error of the learned model, while at the physically more realistic argon coupling its signal remains below this error floor for our data and architecture. This provides a measure of the sensitivity of trajectory-based recovery. 

A particularly important aspect of the successful recovery is the design of the data. For a singular and rapidly varying interaction, the sampling strategy is a central part of the learning problem. The data must adequately represent the relevant radial and angular regimes while avoiding both near-singular configurations and dynamically irrelevant far-field regions. Our experiments showed that these choices directly determine whether the steep radial dependence and angular structure can be learned. The resulting log-uniform, mixture-weighted sampling therefore provides a physically motivated way of allocating data toward the regions where the interaction is most informative, and we expect this tailoring to be a transferable ingredient for learning other singular many-body interactions. 

The same principle extends to the model itself. By fixing the known kinetic term and restricting the unknown potential to depend only on positions, we incorporate established physical structure directly into the learning problem. This makes the approach substantially more interpretable than a fully black-box dynamical model while preserving the flexibility of deep learning. Within these assumptions, the learned Lagrangian also provides a direct representation from which physical observables and other dynamical quantities can be obtained.

Finally, symbolic regression turns the learned representation into an explicit physical law. From trajectory data and network's own output, without providing any information about its functional form, it recovers the analytic structure of the AT interaction and its amplitude, including the $\nu\simeq1$ coupling dependence. This combination of LNN-based dynamical recovery and symbolic inference is, to our knowledge, among the first demonstrations of recovering an explicit irreducible three-body potential in this way. More broadly, it illustrates how deep learning can be used to extract interpretable physical structure from dynamics reproduced via itself.

Taken together, these results establish a framework in which physically structured deep learning can serve as a tool for scientific inference. Known structure can be encoded rather than relearned, challenging interactions can be exposed through appropriate data design, and the resulting learned dynamics can be converted into explicit candidate laws. The Axilrod--Teller interaction provides a difficult and physically meaningful test of this idea, while the methodology offers a route toward studying other systems in which higher-order interactions are difficult to formulate or their structure is not known a priori. Its applicability (without requiring any modifications) beyond the present setting will depend on the validity of assumptions such as a position-only potential and constant mass matrix, but within these boundaries, trajectory data can contain enough information to both reproduce dynamics, \textit{and} recover the physics underlying them.

\section*{Data availability}
The code used to generate the datasets, train and evaluate the models, and produce
the figures, together with the datasets and the trained model weights, is openly
available at \url{https://github.com/senakalabalik/at-lnn} and archived at
Zenodo~\cite{ATrelease}. It builds on and extends the Lagrangian-neural-network
implementation of Ref.~\cite{hamza2026}, available at~\cite{hamzaRepo}.

\begin{acknowledgments}
The authors used Claude (Anthropic), a large language model, as an assistant during this work: to help write and debug analysis and plotting code, to cross-check numerical evaluations, to check the consistency of reported results, and to suggest wording and revisions to the manuscript text. The research questions, physical models, methodology, and interpretation of the results are the authors' own. All AI-assisted code, evaluations, and suggestions were reviewed and verified by the authors, who take full responsibility for the content of this paper.
\end{acknowledgments}

\FloatBarrier
\appendix

\section{Dataset generation}
\label{app:dataset}

Configurations are constructed directly in shape space. For each sample: particle 1 is placed uniformly in the simulation box; particles 2 and 3 are placed at distances $r_{12}$ and $r_{31}$ from it, each with uniformly random orientation; the configuration is accepted if all three pairwise distances lie in $[r_{\min}, r_{\max}]$ and all particles remain in the box, and rejected otherwise. Distances are drawn from a log-uniform mixture: with probability $p_{\mathrm{dense}}$ from $[r_{\min}, r_{\mathrm{dense}}]$ and otherwise from $[r_{\mathrm{dense}}, r_{\max}]$, concentrating samples where forces are large and nonlinear while retaining coverage of the weak-interaction regime. Each accepted configuration receives a random global rotation and translation, and particle labels are randomly permuted, enforcing the exact symmetries of the problem at the data level. Velocities are drawn from an isotropic Gaussian, they are stored as part of the state and used to seed trajectory evaluation, but they enter training only through the fixed kinetic term and Euler--Lagrange algebra, never the potential network (Sec.~\ref{sec:input-rep}).

The datasets used throughout this work are summarized in Table~\ref{tab:dataset_params}.

\begin{table}[h]
\caption{Dataset-generation parameters.}
\label{tab:dataset_params}
\begin{ruledtabular}
\begin{tabular}{lcccc}
System &
$r_{\min}$ &
$r_{\mathrm{dense}}$ &
$r_{\max}$ &
$p_{\mathrm{dense}}$ \\
\colrule
AT & 0.6 & 1.5 & 5.0 & 0.9\\
LJ & 0.9 & 1.5 & 3.0 & 0.5\\
LJ$+$AT & 0.9 & 1.5 & 3.0 & 0.5\\
\end{tabular}
\end{ruledtabular}
\end{table}

Although the configurations are generated synthetically, the protocol does not depend on this. Because $V_\theta$ depends only on positions and the loss is evaluated per sample, any set of position–acceleration pairs can be used, including frames from recorded trajectories with accelerations obtained by finite differences. Such data would first need to be resampled to match the dense/sparse distribution. Trajectories sample configurations in proportion to the time spent in them, and so under-represent the compact, strong-force geometries that Sec.~\ref{sec:experiments} shows are essential. The practical cost would be therefore set by the number of close encounters that can be recorded, and by the potential amplification of position noise in finite-difference accelerations near close approach.

\section{Training on trajectory data}
\label{app:trajectory}

As opposed to the models in the main text which are trained on sampled configurations with exact accelerations, here we also train on frames of
simulated trajectories, with accelerations obtained by finite differences, as they would be from measured positions. The trajectory
models are evaluated on held-out trajectories drawn from the same distribution as their training data. That distribution does not
coincide with the configuration domain of the main results, and we do not claim validity there.

\paragraph{Data.}
We integrate 20\,000 trajectories of the isolated AT system (RK4, $\Delta t = 10^{-3}$, up to $t = 4$). Initial triangles have sides
in $[0.8, 2.5]$, initial velocities are Gaussian with standard deviation $0.3$ and zero total momentum, and a trajectory is stopped
once any separation leaves the training range $[0.6, 5]$. Positions are recorded every $h = 5\times10^{-3}$, and accelerations are
obtained with a Savitzky--Golay filter (window 9, cubic). Against the exact accelerations, these labels have a relative error of
$0.7\,\%$, measured as $\sum\lVert\Delta\ddot{\mathbf q}\rVert / \sum\lVert\ddot{\mathbf q}\rVert$ over all frames.
Free trajectories visit configurations with very different frequencies, so the training frames are drawn with probability proportional
to the ratio of a target histogram of the smallest separation (that of the main-text training set) to the trajectories' own. This is a
reweighting based on positions only, using the physical insight that the force is concentrated at small separations, equivalent to reweighting loss on raw frames.

\paragraph{Training and accuracy.}
The network, features and optimizer are those of the main text.%
We draw 120\,000 frames and train for 200 epochs. For testing, we use 50\,000 frames of 4000 independently generated trajectories.
The relative force error is $2.1\,\%$ against the exact accelerations, and $1.9\,\%$ against the finite-difference labels of the test
frames (Table~\ref{tab:traj}).

\paragraph{Symbolic regression.}
We apply the symbolic regression of Sec.~\ref{sec:exp_at_symbolic}, with the same features, operators, loss and search budget, to the
learned potential on 4000 configurations of the held-out trajectories, reweighted as above. The additive constant of $V_\theta$ is fixed
by setting $V_\theta = 0$ for a widely separated equilateral triangle (side $4.9$). As a control, the same search applied to the exact
AT potential on the same configurations returns $(1 + 3c_1c_2c_3)/(r_{12}r_{23}r_{31})^3$ exactly, in both search seeds, and also with
the reduced operator sets of App.~\ref{app:symbolic} (no cube and binary operators only). On the learned potential, the expression selected
by PySR's 'best' model-selection criterion is
\begin{equation}
V_{\mathrm{SR}} = \frac{1.016 + 3.031\, c_1c_2c_3}{(r_{12}r_{23}r_{31})^3} - 0.020,
\label{eq:traj_sr}
\end{equation}
i.e.\ the AT form with $\nu = 1.016$ and an angular coefficient $3.031/1.016 = 2.98$ (exact values $1$ and $3$), the constant is the
irrelevant additive offset. The second search seed gives the same form ($\nu = 1.018$, coefficient $2.98$) plus a term $6\times10^{-4}\,r_{23}$.
Forces from Eq.~\eqref{eq:traj_sr} agree with the exact AT forces on 3000 further held-out trajectory configurations to within $0.9\,\%$.

\paragraph{Lower-cost checks (60 epochs).}
Three further variants were trained for 60 epochs only.
(i)~Without reweighting (frames as visited), symbolic regression again returns the AT form, $\nu = 1.01$ and $0.99$, angular coefficient
$3.02$ and $3.01$ in the two seeds, with small additional terms.
(ii)~On the reweighted 60-epoch model, the search with binary operators only returns the AT form in both seeds, and without the cube
operator, in one of two seeds.
(iii)~We add Gaussian noise of standard deviation $\sigma$ to the recorded positions before differentiation. Since finite differences
amplify it as $\sigma/h^2$, $\sigma = 10^{-5}$ and $3\times10^{-5}$ give labels with $27\,\%$ and $78\,\%$ relative error. The network
still reaches $5.8\,\%$ and $7.6\,\%$ on the held-out trajectories, and symbolic regression returns the AT form in one of two search seeds
at each level ($\nu = 0.99$, coefficient $3.04$; and $\nu = 0.97$, coefficient $3.11$, with small additional terms). Larger position noise
might require a coarser frame interval or a noise-robust differentiator. 

\begin{table}[h]
\caption{Relative acceleration error on held-out trajectories (50\,000 frames from 4000 independent trajectories), against exact and against
finite-difference (FD) accelerations. Training data: trajectory frames with FD labels.}
\label{tab:traj}
\begin{ruledtabular}
\begin{tabular}{lcc}
model & exact & FD labels \\
\hline
reweighted, 200 epochs & 0.021 & 0.019 \\
reweighted, 60 epochs & 0.031 & 0.030 \\
as visited, 60 epochs & 0.032 & 0.034 \\
reweighted, 60 epochs, $\sigma = 10^{-5}$ & 0.058 & 0.054 \\
reweighted, 60 epochs, $\sigma = 3\times10^{-5}$ & 0.076 & 0.072 \\
\hline
FD labels themselves (noise-free) & 0.007 & --- \\
\end{tabular}
\end{ruledtabular}
\end{table}

\section{Training details}
\label{app:training}

Training details are summarized in the table \ref{tab:training_params}.

\begin{table}[h]
\caption{Training hyperparameters used throughout this work.}
\label{tab:training_params}
\begin{ruledtabular}
\begin{tabular}{lc}
Hyperparameter & Value \\
\colrule
Optimizer & RAdam \\
Learning rate & $10^{-3}$ \\
Weight decay & $10^{-6}$ \\
Learning-rate scheduler & Cosine annealing \\
Scheduler period ($T_{\max}$) & 20 epochs \\
Minimum learning rate ($\eta_{\min}$) & $5\times10^{-4}$ \\
Batch size & 128 \\
Training epochs & 200 \\
Hidden layers & 3 \\
Hidden width & 256 \\
Activation & GELU \\
\end{tabular}
\end{ruledtabular}
\end{table}

\section{Evaluation metrics}
\label{app:eval_metrics}

Throughout the Experiments section we report the standard mean relative error
(MRE), $\mathrm{MRE}=\tfrac{1}{n}\sum_i |\hat x_i-x_i|/|x_i|$ for a quantity $x$
(potential or acceleration magnitude) at $n$ held-out points. This is a mean of
\emph{pointwise} ratios, and it is unstable wherever $x_i$ passes near zero.
A single point with a small true value can dominate the average regardless of
how accurate the fit is everywhere else. This happens for physical
reasons in our data. The Axilrod--Teller angular factor $\Phi$ changes sign,
the Lennard--Jones potential crosses zero at $r=\sigma$, and any three-body
contribution vanishes as a particle recedes, so a handful of unremarkable
points can otherwise produce a nonsensical headline error.

Wherever this occurs we instead report the magnitude-weighted \emph{relative L1
error},
\begin{equation}
    \varepsilon_x = \frac{\sum_i |\hat x_i - x_i|}{\sum_i |x_i|},
    \label{eq:rel_l1}
\end{equation}
which aggregates errors and magnitudes separately before dividing once. Writing
$w_i = |x_i|/\sum_j|x_j|$, this is exactly the magnitude-weighted average
$\varepsilon_x=\sum_i w_i\,(|\hat x_i-x_i|/|x_i|)$ of the same pointwise ratios
MRE averages unweighted. Points with a small true value are down-weighted
automatically. The two coincide only when every $|x_i|$
is equal, and can diverge sharply otherwise. Table~\ref{tab:mre_l1_at} shows
both, side by side, for the two evaluations in this work where the difference is
largest. Table~\ref{tab:mre_l1_lj} gives the same comparison for the pure-LJ control.

\begin{table}[h]
\caption{Radial dependence (pure AT, $r\in[r_{\min},1.6]$): MRE vs.\ relative L1
error, per reference shape. The far-particle family's net force vanishes fastest
as the third atom recedes, so a small absolute residual there becomes a large
\emph{relative} one under either metric, reported per-shape.}
\label{tab:mre_l1_at}
\begin{ruledtabular}
\begin{tabular}{lcccc}
Shape & MRE ($V$) & L1 ($V$) & MRE ($a$) & L1 ($a$) \\
\colrule
Equilateral   & 0.29\%   & 0.32\%  & 3.28\%    & 1.78\% \\
Right         & 0.42\%   & 0.18\%  & 5.73\%    & 0.27\% \\
Line          & 0.33\%   & 0.22\%  & 10.77\%   & 0.88\% \\
Far particle  & 194.59\% & 87.64\% & 8764.27\% & 21.59\% \\
Isosceles     & 0.56\%   & 0.22\%  & 11.20\%   & 0.20\% \\
Scalene       & 0.74\%   & 0.33\%  & 11.62\%   & 0.17\% \\
\colrule
All 6 (agg.)      & 32.82\% & 14.82\% & 1467.81\% & 4.15\% \\
Excl.\ far particle & 0.47\%  & 0.25\%  & 8.52\%    & 0.66\% \\
\end{tabular}
\end{ruledtabular}
\end{table}

\begin{table}[h]
\caption{Same comparison for the pure-LJ control ($r\in[0.9,1.6]$). The far-particle divergence between MRE and L1
is far milder here than for AT (Table~\ref{tab:mre_l1_at}). Direct numerical
evidence of the $r^{-6}$ vs.\ $r^{-9}$ decay-rate difference between the two
interactions.}
\label{tab:mre_l1_lj}
\begin{ruledtabular}
\begin{tabular}{lcccc}
Shape & MRE ($V$) & L1 ($V$) & MRE ($a$) & L1 ($a$) \\
\colrule
Equilateral   & 0.55\% & 0.53\% & 1.54\% & 1.23\% \\
Right         & 1.17\% & 0.16\% & 0.34\% & 0.48\% \\
Line          & 0.99\% & 0.52\% & 0.84\% & 0.99\% \\
Far particle  & 1.56\% & 0.77\% & 7.39\% & 0.96\% \\
Isosceles     & 1.51\% & 0.46\% & 0.46\% & 0.59\% \\
Scalene       & 0.66\% & 0.27\% & 0.25\% & 0.33\% \\
\colrule
All 6 (agg.)      & 1.07\% & 0.45\% & 1.81\% & 0.76\% \\
Excl.\ far particle & 0.98\% & 0.39\% & 0.69\% & 0.72\% \\
\end{tabular}
\end{ruledtabular}
\end{table}

Where a single fit-quality number is
more natural than a paired potential/acceleration error, such as the Newtonian control and
the angular-factor recovery diagnostic, we instead report the coefficient of
determination,
\begin{equation}
    R^2 = 1 - \frac{\sum_i (x_i-\hat x_i)^2}{\sum_i (x_i-\bar x)^2},
\end{equation}
with $\bar x$ the mean of $x_i$ over the evaluation set. $R^2=1$ is a perfect fit and
$R^2=0$ is no better than predicting the mean.

\section{Robustness of the symbolic recovery}
\label{app:symbolic}

This appendix supports the claims of Sec.~\ref{sec:exp_at_symbolic} with the
checks omitted from the main text for space.

\subsection{Operator-set dependence}

A natural objection to Eq.~\eqref{eq:sr-recovered} is that the operator set
available to the search, which included a cube and a reciprocal unary operator,
was chosen with $(r_{12}r_{23}r_{31})^{-3}$ in mind and so predisposed the search
toward $\AT$. We test this directly on $V_\theta$ by repeating the search (i)
with the cube operator removed, leaving only a square and a reciprocal, and (ii)
with \emph{no} unary operators at all, so that any power must be built by
repeated binary multiplication. Both variants recover the same angular numerator
and radial scaling as Eq.~\eqref{eq:sr-recovered} ($R^2\approx0.95$ against the
analytic $\AT$ in both cases, against $R^2\approx1.0$ fitting $V_\theta$ itself),
confirming that the recovered structure reflects $V_\theta$ and not the
operator budget. On the analytic control, the same two variants, together with a
broadened set including an exponential and a sine as unrelated distractor
operators, all recover Eq.~\eqref{eq:AT} to $R^2\gtrsim0.99$.

\subsection{Non-uniqueness.} The angle cosines are not independent of the side
lengths (Sec.~\ref{sec:input-rep}): for any triangle,
$\cos^2\theta_1+\cos^2\theta_2+\cos^2\theta_3+2\cos\theta_1\cos\theta_2\cos\theta_3=1$.
Consequently Eq.~\eqref{eq:sr-recovered} is not obligated to be the unique expression matching
$V_\theta$: any two polynomials in $\cos\theta_i$ differing by a multiple of this
identity coincide on every physical triangle. Different search runs accordingly may
return algebraically distinct but numerically equivalent representatives (e.g.\
an expression in $\cos\theta_1\cos\theta_2\cos\theta_3$ versus one in
$\cos^2\theta_1+\cos^2\theta_2+\cos^2\theta_3$). Hence we can verify their equivalence by reducing both expressions to a common canonical form using the identity.


\end{document}